\documentclass[acmtog, authorversion]{acmart}

\usepackage{enumitem}

\AtBeginDocument{%
  \providecommand\BibTeX{{%
    \normalfont B\kern-0.5em{\scshape i\kern-0.25em b}\kern-0.8em\TeX}}}

\setcopyright{acmcopyright}
\copyrightyear{2018}
\acmYear{2018}
\acmDOI{XXXXXXX.XXXXXXX}

\acmConference[Conference acronym 'XX]{Make sure to enter the correct
  conference title from your rights confirmation emai}{June 03--05,
  2018}{Woodstock, NY}
\acmBooktitle{Woodstock '18: ACM Symposium on Neural Gaze Detection,
 June 03--05, 2018, Woodstock, NY} 
\acmPrice{15.00}
\acmISBN{978-1-4503-XXXX-X/18/06}

\begin{document}

\title{Building a Llama2-finetuned LLM for Odia Language Utilizing Domain Knowledge Instruction Set}




\author{\large{Guneet Singh Kohli}}
\affiliation{%
  \institution{Thapar University}
  \country{India}}

\author{\large{Shantipriya Parida}}
\affiliation{%
  \institution{Silo AI}
  \country{Finland}
}
\email{shantipriya.parida@silo.ai}

\author{\large{Sambit Shekhar}}
\affiliation{%
 \institution{Odia Generative AI}
 \country{India}}



 \author{\large{Samirit Saha}}
 \authornote{Both authors contributed equally to this research.}
\author{\large{Nipun B Nair}}
\authornotemark[1]
 \affiliation{%
   \institution{Amrita Vishwa Vidyapeetham}
  \country{India}
}

\author{\large{Parul Agarwal}}
\affiliation{%
  \institution{Institute of Mathematics and Applications}
  \country{India}}

\author{\large{Sonal Khosla}}
\affiliation{%
  \institution{Odia Generative AI}
  \country{India}
  }

\author{\large{Kusumlata Patiyal}}
\affiliation{%
  \institution{Sharda University}
  \country{India}
  }

\author{\large{Debasish Dhal}}
\affiliation{%
  \institution{NISER Bhubaneswar}
  \country{India}}
  
\renewcommand{\shortauthors}{Kohli, et al.}

\begin{abstract}
Building LLMs for languages other than English is in great demand due to the unavailability and performance of multilingual LLMs, such as understanding the local context. The problem is critical for low-resource languages due to the need for instruction sets. In a multilingual country like India, there is a need for LLMs supporting Indic languages to provide generative AI and LLM-based technologies and services to its citizens.  

This paper presents our approach of i) generating a large Odia instruction set, including domain knowledge data suitable for LLM fine-tuning, and ii) building a Llama2-finetuned model tailored for enhanced performance in the Odia domain. 

The proposed work will help researchers build an instruction set and LLM, particularly for Indic languages. We will release the model and instruction set for the public for research and noncommercial purposes. 
\end{abstract}




\maketitle

\section{Introduction}
In the ever-evolving landscape of Generative AI, language models have played a pivotal role in revolutionizing how we interact with and understand the world's diverse linguistic tapestry. These models have demonstrated remarkable capabilities, from the groundbreaking BERT \cite{devlin2018bert} to the omnipotent GPT, particularly in widely spoken languages. However, the realm of Generative AI is not one-size-fits-all, and the inadequacy of state-of-the-art models when applied to less-resourced languages, especially those belonging to the Indic language family, remains an unsettling challenge.

The multilingual landscape of NLP has witnessed significant advancements in recent years, with several models tailored to support a myriad of languages. However, a noticeable gap in the digital representation persists when it comes to providing comprehensive support for Indic languages, including Odia. This gap fuels our dedication to the cause – empowering the Odia language through robust and dedicated language models.

\begin{figure}[!htb]
\begin{center}
\includegraphics[scale=0.19]{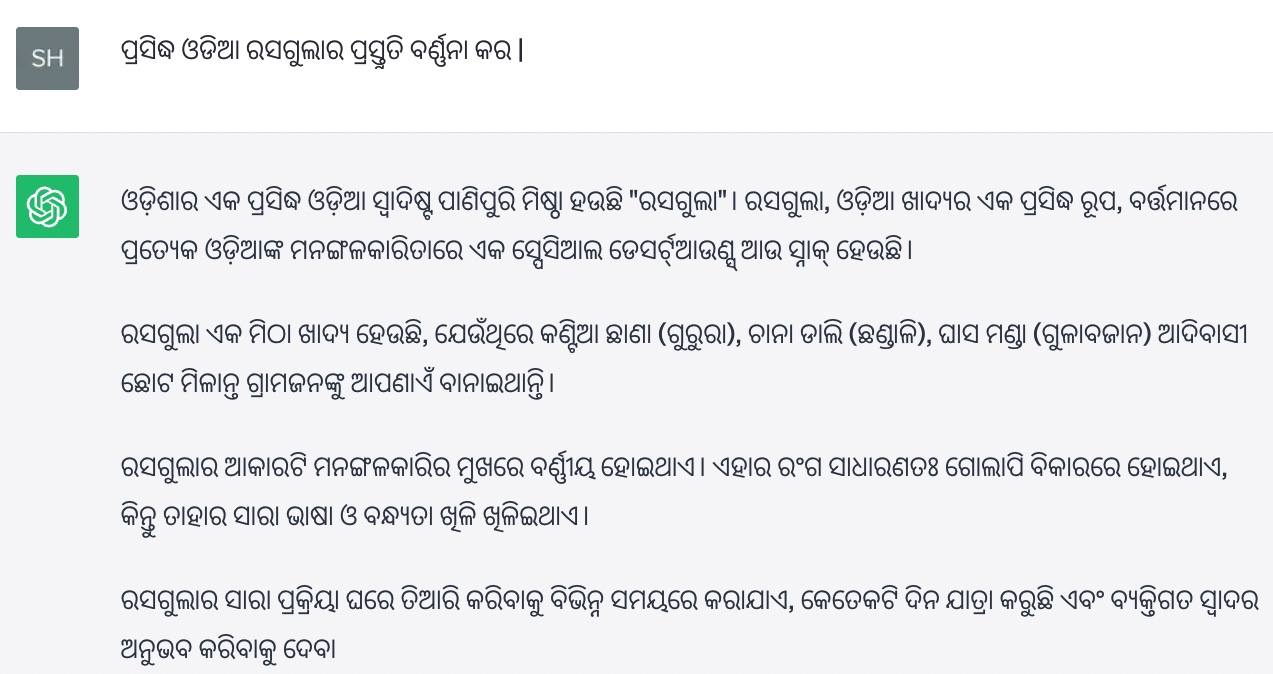} 
\caption{The queries to ChatGPT3.5 about preparing the famous Odia sweet: Rasgulla. The answer from ChatGPT3.5 is not relevant and noisy.}
\label{fig_chatgpt}
\end{center}
\end{figure}

Odia, an Indic language spoken predominantly in the Indian state of Odisha, boasts a rich cultural heritage and a growing digital presence. However, it has long been underserved in the digital sphere, and the need to empower it through enhanced language representation has never been more pressing \cite{parida2022universal}. Although only some multilingual LLMs support the Odia language due to the amount and quality of data, the performance of these LLMs could be better, as shown in Fig. \ref{fig_chatgpt}.    

In this paper, we embark on a journey to introduce the Llama2-finetuned language model meticulously tuned to cater to the specific requirements of the Odia language. Llama2 \cite{touvron2023llama} proudly stands as a member of a burgeoning family of models, poised to bridge the multilingual gap by providing precise, contextually relevant results for Odia while championing linguistic diversity in the NLP landscape.

Our motivation transcends the boundaries of linguistic capabilities; it is rooted in the broader mission of preserving and promoting underrepresented languages. Llama2's development and meticulous fine-tuning of an extensive Odia domain knowledge dataset serve not only to enhance its performance but also to contribute to safeguarding against potential plagiarism in Odia language research. The authors aspire to foster innovation and knowledge creation within the Odia-speaking community by offering researchers a dependable and original language model.

\section{DATASET}
In the pursuit of empowering the Odia language in the digital sphere, it is recognized by the team that data forms the bedrock of progress. To this end, a comprehensive dataset of Llama2, the specialized language model designed exclusively for Odia, has been meticulously curated and prepared. The dataset encapsulates a wealth of linguistic diversity, domain-specific knowledge, and an expansive coverage of the Odia language landscape. The dataset is a versatile repository that encompasses two distinct types of data.
\paragraph{Translated Instruction Set:}This dataset segment is a testament to the team's commitment to linguistic diversity. A curated collection of translated instructions covering a wide range of topics and contexts is encompassed by it. The data was obtained from various internet sources, primarily blogs and Wikipedia. They were translated using the Indic Trans library under the supervision of colloquial speakers of the Odia language, who edited any discrepancies manually. These instructions serve as valuable training data for Llama2, enabling it to understand and generate Odia text that remains faithful to the nuances and subtleties of the language. 
\paragraph{Domain Knowledge:}The necessary expertise to navigate Odia's unique linguistic intricacies has been equipped to Llama2 by incorporating domain knowledge into the dataset. This knowledge is wide-ranging and deeply rooted in Odia's culture, history, and contemporary context. The power of sources such as Wikipedia and ChatGPT has been harnessed to provide Llama2 with a rich understanding of Odia, enhancing its capacity to generate contextually relevant and culturally sensitive content. 

The instructions cover a wide array of subjects, ranging from the culinary delights in \textit{RECIPES}, the historical significance of \textit{HISTORICAL PLACES}, and \textit{TEMPLES OF ODISHA}, to the intellectual pursuits in \textit{ARITHMETIC}, \textit{HEALTH}, and \textit{GEOGRAPHY}. It also explores the artistic tapestry of Odisha through \textit{ART AND CULTURE}, which celebrates renowned figures in \textit{FAMOUS ODIA POETS/WRITERS}, and \textit{FAMOUS ODIA POLITICAL LEADERS}. Furthermore, it encapsulates \textit{SPORTS} and the \textit{GENERAL KNOWLEDGE OF ODISHA}, providing an all-encompassing representation of the state. These instructions reflect Odisha's rich heritage and are a practical and engaging resource for building a conversational AI that resonates with the region's people.

The dataset contains 181K Odia instruction sets \cite{odia_master_data_llama2}. The instruction sets include:

 \begin{itemize}[noitemsep]
    \item Odia domain knowledge instruction set prepared using various domain knowledge sources utilizing GPT models.
    \item Translating popular instruction sets (\textit{Alpaca} \cite{taori2023alpaca}, \textit{Dolly\footnote{\url{https://huggingface.co/datasets/databricks/databricks-dolly-15k}}}, \textit{GPT Teacher \footnote{\url{https://github.com/teknium1/GPTeacher}}}) from English to Odia using the IndicTrans\footnote{\url{https://github.com/AI4Bharat/indicTrans}} machine translation library from AI4Bharat \cite{indic_trans}. 
    \item Preparing translation instruction set using OdiEnCorp \cite{parida-etal-2020-odiencorp}, English-Odia parallel corpus.
    \item Hard-coded instruction set. 
\end{itemize}

The statistics of the dataset are shown in Table \ref{tab:fine_tune_parameter}. 

\begin{table}[!htb]
\centering
\small
\begin{tabular}{|l|l|}
\hline
\bf \bf Dataset & \bf Size \\
\hline
Odia Domain\_Context & 10.5k \\ \hline 
Alpaca & 52k \\ \hline
Dolly & 15k\\ \hline
GPT teacher role-play & 3k \\ \hline 
GPT teacher instruct & 18k \\ \hline
Hard code Q{\&}A & 105 \\ \hline
\end{tabular}
\caption{Details of the data used in the instruction fine-tuning stage.}
\label{tab:fine_tune_parameter}
\end{table}

\section{EXPERIMENTAL SETUP}
\subsection{Fine-Tuning}
We used the Llama2-7b \footnote{\url{https://huggingface.co/meta-llama/Llama-2-7b}} which has 4k context length as our base model for fine-tuning. 
A deliberate and balanced approach was undertaken to optimize efficiency and performance in configuring our model's parameters. Employing a batch size of 128 strikes a harmonious equilibrium, allowing the processing of 128 data samples concurrently while efficiently managing memory resources. A learning rate of $2e^{-4}$ was selected, signifying an informed choice that facilitates model convergence without the risk of overshooting \cite{lewkowycz2021decay}. The decision to train the model for seven epochs harnesses the power of iterative learning, enabling it to refine its understanding of the data progressively. To ensure streamlined processing, we imposed a cutoff length of 256, which efficiently manages computational resources for handling input sequences. A weight decay factor of 0.001 strikes a careful balance, enhancing generalization without imposing undue constraints on the model's flexibility \cite{ginsburg2019stochastic}. A warmup ratio of 0.03 in the learning rate schedule adds training stability by gradually increasing the learning rate during early iterations.

Setting a maximum sequence length of 512 accommodates the complexities of capturing context in NLP tasks. Adopting a linear learning rate scheduler initially promotes faster convergence with a higher learning rate \cite{wang2022incremental}. Fine-tuning attention mechanisms with LoRA (Learnable Reordering Attention) parameters \cite{hu2021lora}, such as LoRA r (64) and LoRA $\alpha$ (16), refines the model's attention operations. A dropout rate of 0.1 introduces a measured dropout level, mitigating overfitting risks. Leveraging 4-bit quantization optimizes memory and computational efficiency without compromising performance \cite{wei2022outlier}. Setting the compute data type to ``float16" further balances precision and computational speed. While nested quantization, although not utilized in this instance, represents a potential avenue for additional efficiency gains. Batch size 48 for training and evaluation meticulously manages memory and training stability \cite{dai2022reveal}. The accumulation of gradients over a single update step enhances stability without necessitating additional memory. 

A maximum gradient norm of 0.3 safeguards against unstable training by preventing gradient explosions \cite{arras2016explaining}. Employing the ``Paged AdamW 32-bit" optimizer signifies a specialized adaptation for 32-bit computations, enhancing training efficiency \cite{dettmers20218}. The selection of a constant learning rate schedule ensures steady learning rate values conducive to convergence. Lastly, attention operations are finely tuned by targeting specific projection modules (q\_proj, k\_proj, v\_proj, o\_proj) with LoRA attention, potentially elevating performance in those particular model areas. This holistic parameter configuration approach harmonizes computational resource management and model effectiveness in our research endeavor. 

\subsection{TRAINING}

The training environment has been trained on NVIDIA A-100 PCI Express, 40 GB single core because PCI Express boasts exceptional computational power, enabling it to handle the complex mathematical operations required during the training of deep neural networks, such as the Llama2 model \cite{fang2022parallel}. Its advanced architecture allows for rapid parallel processing, significantly accelerating model training. The model completed its first training in approx 2.5 days. The training hyperparameters are shown in Table \ref{tab:train_param}.   

\begin{table}[!htb]
\centering
\small
\begin{tabular}{|l|l|}
\hline
\bf \bf Hyper Parameter & \bf Value \\
\hline
Batch Size & 128 \\ \hline
Learning Rate & $2e^{-4}$\\ \hline
Epochs        & 7 \\ \hline 
Cutoff Length & 256 \\ \hline 
Weight\_Decay & 0.001 \\ \hline
Warmup\_Ratio & 0.03 \\ \hline
max\_seq\_length & 512 \\ \hline
LR\_Scheduler & linear \\ \hline
Lora r & 64 \\ \hline
Lora $\alpha$ & 16 \\ \hline
Lora dropout & 0.1 \\ \hline
use\_4bit & True \\ \hline
bnb\_4bit\_compute\_dtype & ``float16" \\ \hline
bnb\_4bit\_quant\_type & ``nf4" \\ \hline
use\_nested\_quant & False \\ \hline
per\_device\_train\_batch\_size & 48 \\ \hline
per\_device\_eval\_batch\_size & 48 \\ \hline
gradient\_accumulation\_steps & 1 \\ \hline
max\_grad\_norm & 0.3 \\ \hline
optim & ``paged\_adamw\_32bit" \\ \hline
lr\_scheduler\_type & ``constant" \\ \hline
Lora Target Modules & (q\_proj, k\_proj, v\_proj, o\_proj) \\ \hline
\end{tabular}
\caption{Training Hyperparameters.}
\label{tab:train_param}
\end{table}

\begin{figure}[!htb]
\centering
\includegraphics[scale=0.20]{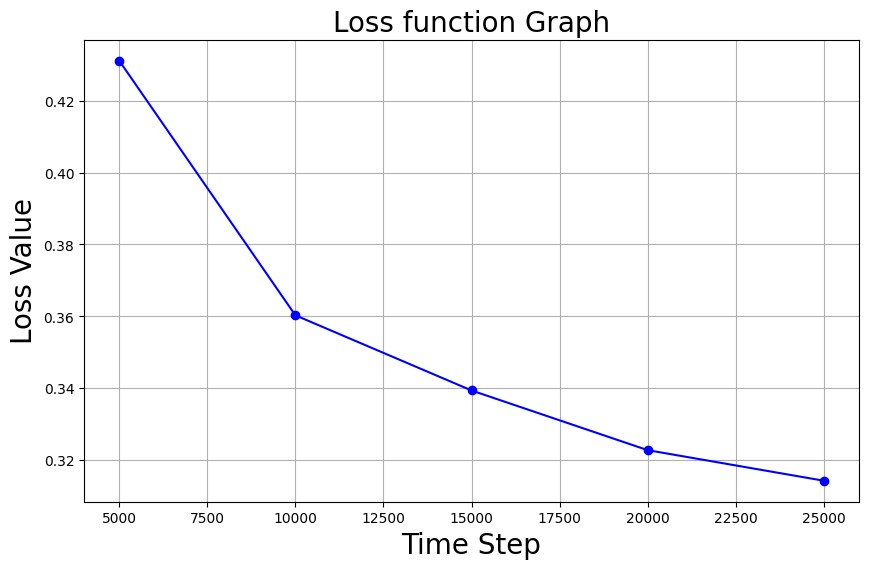} 
\caption{The Loss graph of the model}
\label{lossgraph}
\end{figure}

We visually represent our model's learning process in the training section through a loss function graph. The graph showcases the progression of time steps along the x-axis and the corresponding loss values on the y-axis. What becomes immediately apparent is the model's steady improvement over time, as indicated by the decreasing loss values \cite{viering2022shape}. This trend signifies the model's ability to learn and adapt as it refines its task understanding. The data points, spanning key milestones at 5000, 10000, 15000, 20000, and 25000 time steps, offer concrete evidence of this learning trajectory. The decreasing loss values at these checkpoints, starting at 0.4311 and culminating at 0.3143, underscore the model's continuous refinement, validating its training effectiveness and the progress made throughout the training process in Fig. \ref{lossgraph}.

\section{INFERENCE}

The inference script is adapted from Alpaca-LoRA considering the base model Llama2-7b with the built model weights. 

\subsection{Text Generation Setup}
The decoding process of LLMs plays a critical role in determining the quality and diversity of the generated text. In our experiments, we use the following decoding hyperparameters:


\begin{itemize}[noitemsep]

    \item \textit{Size of the context}: We establish the context size as 2048, determining the maximum number of tokens that the model can take into account simultaneously during the text generation process. 
    \item \textit{Maximum sequence length}: We impose a constraint on the generated sequence length, limiting it to 512 tokens to ensure that the outputs remain focused and closely related to the input prompt.
    
    \item \textit{Temperature}: We set the temperature to 0.2, regulating the level of randomness in the sampling process. Lower values make the model produce more focused and deterministic outputs, while higher values introduce greater diversity at the expense of coherence.
    
    \item \textit{Top-k sampling}: For each step, we adopt Top-k sampling with a value of k = 40, whereby the model selects the subsequent token from the top 40 most probable options. This introduces an element of randomness and diversity in the generated text.
        
    \item \textit{Top-p sampling}: We also employ Top-p sampling with a value of p = 0.9. This further enhances diversity by considering a dynamic set of tokens that collectively account for 90\% of the probability mass.
        
     \item \textit{Repetition penalty}: To discourage the model from generating repetitive text, we apply a repetition penalty with a factor of 1.3, penalizing the selection of already chosen tokens. 
\end{itemize}

\section{EVALUATION}

Evaluation metrics like ROUGE and BLEU were used to assess the quality and performance of the model. ROUGE is a set of metrics commonly used for evaluating the quality of text summaries \cite{yashaswini2021metrics}. It measures the overlap between the words or n-grams (sequences of words) in the generated text and the reference text. ROUGE is widely used to evaluate machine-generated summaries, machine translation, and text-generation tasks. BLEU is designed to assess the adequacy of translations by comparing them to human-generated reference translations. BLEU is a standard metric in machine translation evaluation \cite{saadany2021bleu}. We used 280 samples to calculate the BLEU score and the ROUGE score. The BLEU score was 0.6158, and the ROUGE score was 0.6583. The evaluation scores are shown in Table \ref{tab:eval_score}.   

\begin{table}[!htb]
\centering
\small
\begin{tabular}{|l|l|}
\hline
\bf \bf Evaluation metric & \bf Score \\
\hline
ROUGE & 0.6583 \\ \hline
BLEU        & 0.6158 \\ \hline 

\end{tabular}
\caption{\label{model_conf} Automatic Evaluation Scores.}
\label{tab:eval_score}
\end{table}

\subsection{HUMAN EVALUATION}

The human evaluation process is a crucial and multifaceted element in assessing the Odia generative model's performance, adhering to stringent ethical guidelines and user safety concerns. Trained evaluators, possessing expertise in linguistics and a profound understanding of the Odia language, play pivotal roles in this assessment.

\begin{figure}[!htb]
\centering
\includegraphics[scale=0.3]{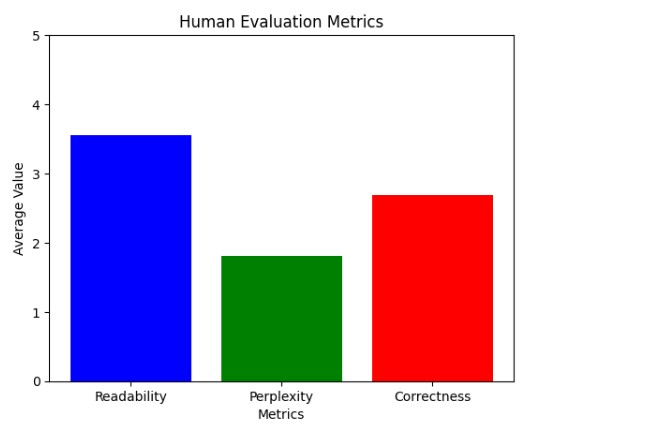} 
\caption{\centering{The score for human evaluation metrics}}
\label{humaneval}
\end{figure}

Apart from automatic evaluation, we performed a human assessment of the model by the native Odia speakers by asking subjective questions to verify model generative performance and toxicity. We did our analysis on the basis of three metrics: readability, perplexity, and correctness. A graphical analysis of the average is presented in Fig. \ref{humaneval}.   

Three critical metrics often take center stage in text analysis and natural language processing: Readability, Perplexity, and Correctness. Readability, a measure of how easily a text can be comprehended by its intended audience, is pivotal in ensuring clarity and accessibility in written communication \cite{mendoncca2023simple}. On the other hand, Perplexity is a crucial gauge for assessing the quality of language models, quantifying their predictive accuracy and understanding of language patterns. Lower perplexity values indicate more proficient models. Lastly, Correctness evaluates the accuracy and fidelity of information within the textual content, measuring alignment with factual accuracy and adherence to linguistic rules \cite{madaan2023self}. These metrics, collectively, empower professionals in fields such as journalism, linguistics, and artificial intelligence to enhance the quality, and reliability of textual content, ultimately advancing the capabilities of language models and text-based applications \cite{chiang2023can}.

\section{ANALYSIS AND DISCUSSION}


Our model demonstrated strong performance in arithmetic-related questions, showcasing its ability to provide accurate and concise answers in this domain. The model performs quite well at toxicity, in fact, it prefers using the formal honorific. The model has a good understanding of the grammar of the Odia language and it is able to identify grammatical elements from a given piece of text. However, in the context of classification tasks, we observed limitations. It struggled to effectively distinguish between different classes and often defaulted to providing explanations of the question rather than assigning classifications. 

Furthermore, one noteworthy limitation we encountered was related to the response length. When tasked with generating long answers or explanations, the model faced challenges due to constraints imposed by the maximum token size. This limitation impacted the model's ability to provide comprehensive and detailed responses in scenarios where lengthy explanations were required.

These observations underscore the need for further research and potential model enhancements to address these specific limitations, particularly in the context of classification tasks and extended responses.

\section{LIMITATION}

Based on the evaluation, we observed the below issues and limitations with the model.

\begin{itemize}
    \item The model still suffers from hallucinations, particularly for long answers.
    Item Still needs to be able to follow arithmetic problems and critical reasoning.
    \item Still answers incorrectly sometimes for the questions related to Odisha, although the model is explicitly trained, including the domain knowledge instruction set.
    \item The model often produces additional unnecessary text, after producing the correct answer. The model is not performing well when it comes to generalization and summarizing.
    \item Under certain circumstances, the model produces insufficient or no output at all. This is especially true when it comes to arithmetic or when the response is supposed to be big.
\end{itemize}

\section{CONCLUSION}
This paper presented a Llama2-finetuned LLM for the low-resource Odia language using a prepared instruction set containing domain knowledge. Several parameters and techniques were employed to optimize the model's efficiency and performance. These included batch size selection, learning rate, training epochs, sequence length, weight decay, warmup ratio, attention mechanisms with LoRA parameters, dropout rate, quantization, gradient norm, optimizer choice, and specific projection module targeting. This comprehensive parameter configuration balanced computational resource management and model effectiveness for the research endeavor.
The future work includes \textit{i)} investigate model limitations \textit{ii)} to perform an in-depth comparison study of the proposed model with the available multilingual LLMs supporting Odia, \textit{iii)} Exploring distilling step-by-step approach \cite{hsieh2023distilling} for build smaller model for comparative analysis, \textit{iv)} Release pre-train Odia LLM model following BLOOM \cite{scao2022bloom} specification.     

\section*{Ethics Statement}
\small

We do not envisage any ethical concerns. The dataset does not contain any personal or personally identifiable information, the source data is already open source, and there are no risks or harm associated with its usage.

\section*{Acknowledgements}
\small

We are thankful to Silo AI, Helsinki, Finland, and Odia Generative AI, Bhubaneswar, India for the necessary support. 

\bibliographystyle{ACM-Reference-Format}
\bibliography{biblio}

\section{appendix}

\subsection{Comparative Analysis}

\begin{figure*}
    \includegraphics[width=0.85\textwidth]{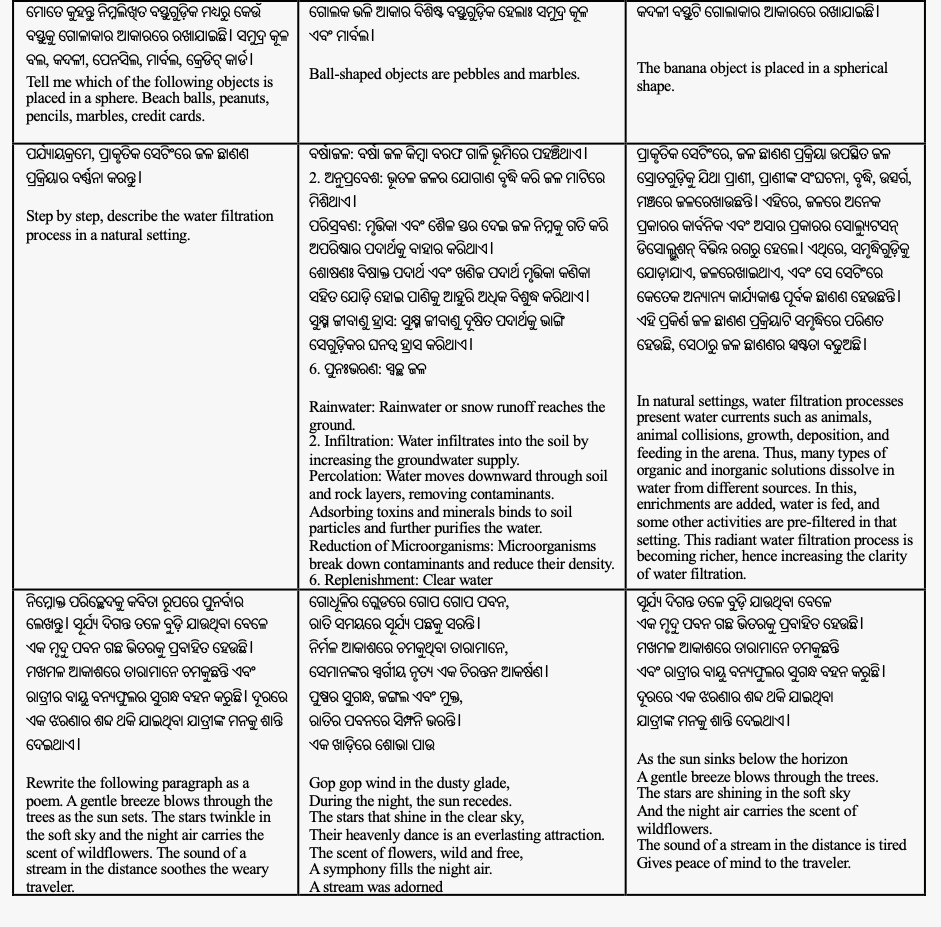}
    \caption{The comparative analysis between our model and ChatGPT 3.5.}
     \label{comp-2}

\end{figure*}

In our study, we conducted a comprehensive comparison between our model and ChatGPT 3.5. The results of this comparative analysis are presented in Figure \ref{comp-2}, which encapsulates the key findings of our research. This comparison serves as a visual representation of the performance and capabilities of our model in relation to ChatGPT 3.5.

\begin{figure*}
    \includegraphics[width=0.85\textwidth]{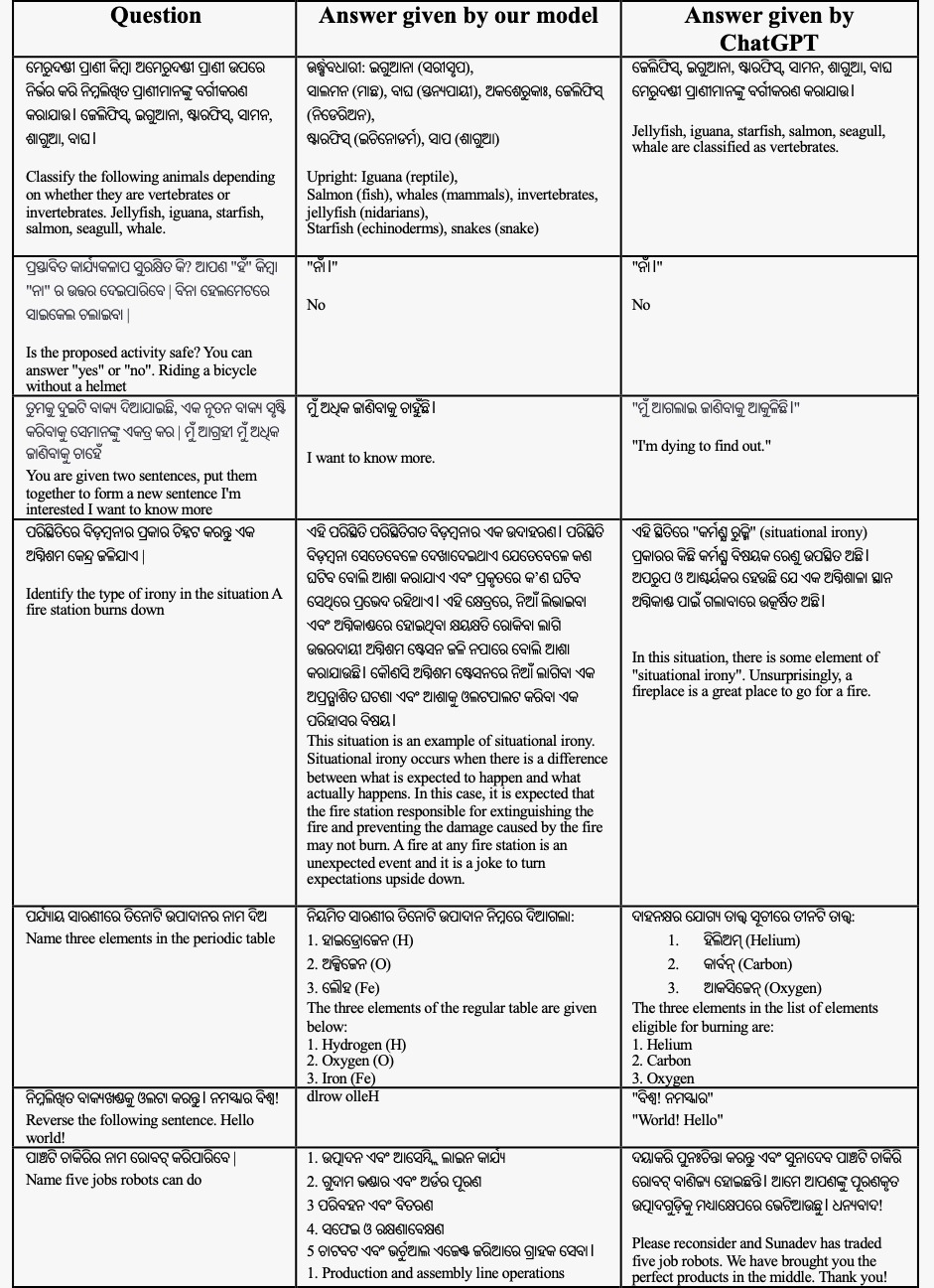}
\end{figure*}

\end{document}